\documentclass[10pt,twocolumn,letterpaper]{article}
\pdfoutput=1
\usepackage{cvpr}
\usepackage{times}
\usepackage{graphicx}
\usepackage{pstool}
\usepackage{amsmath}
\usepackage{amssymb}
\usepackage{cite}
\usepackage{subfigure}
\usepackage{enumitem}
\usepackage{array}
\usepackage{comment}

\usepackage[breaklinks=true,bookmarks=false]{hyperref}

\cvprfinalcopy 

\def\cvprPaperID{****} 

\begin{document}

\title{Deep Visual City Recognition Visualization }

\author{Xiangwei Shi, Seyran Khademi, Jan van Gemert\\
PRB, Computer Vision lab \\
Delft University of Technology\\
}

\maketitle

\begin{abstract}
   Understanding how cities visually differ from each others is interesting for planners, residents, and historians. We investigate the interpretation of  deep features learned by convolutional neural networks (CNNs) for city recognition. Given a trained city recognition network, we first generate weighted masks using the known Grad-CAM technique and to select  the most discriminate regions in the image. Since the image classification label is the city name, it contains no information of objects that are class-discriminate, we investigate the interpretability of deep representations with two methods. (i) Unsupervised method is used to cluster the objects appearing in the visual explanations. (ii) A pretrained semantic segmentation model is used to label objects in pixel level, and then we introduce statistical measures  to  quantitatively evaluate the interpretability of discriminate objects. The influence of network architectures and random initializations in training, is studied on the interpretability of CNN features for city recognition. The results suggest that network architectures would affect the interpretability of learned visual representations greater than different initializations.
\end{abstract}


\vspace{-2ex}
\section{Introduction}
Understanding how cities visually differ from each others is interesting for planners, residents, and historians. Automatic visual recognition is now making great progress which can help identifying how cities visually differ. Creating interpretable convolutional neural network (CNN) is a fascinating path that may lead us towards trustworthy AI~\cite{fong2017interpretable,selvaraju2017grad,simonyan2013deep,springenberg2014striving,zeiler2014visualizing,zhou2014object,zhou2016learning}. Understanding  CNN filters  provides us with valuable insight on decision making criteria for a specific task. Visual features such as objects and parts are examples of high-level semantics that are consistent with  how humans understand and analyze images \cite{bau2017network,li2010object,zhang2018interpretable}. Accordingly, we investigate and evaluate the interpretability of learned discriminate objects in city recognition CNNs.

Visualization of CNN filters are a popular techniques for analyzing CNNs. In this work, we build on top of gradient-weighted class activation mapping (Grad-CAM) method \cite{selvaraju2017grad} to generate class-discriminate visualizations, for our city recognition CNNs. Grad-CAM generates visualizations on the input images with highlight of discriminate regions by analyzing learned convolutional features and taking the information of the fully connected layers into consideration. Grad-CAM does not need to alter structure of the trained CNNs and is model-agnostic.

\begin{figure}
    \centering
    \vspace{-1ex}
    \subfigure[cat and dog image and visualizations]{
    \begin{minipage}[t]{0.25\linewidth}
    \centering
    \includegraphics[width=0.8in]{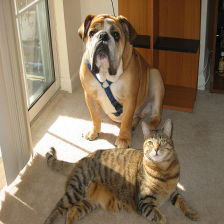}
    \end{minipage}
    \begin{minipage}[t]{0.25\linewidth}
    \centering
    \includegraphics[width=0.8in]{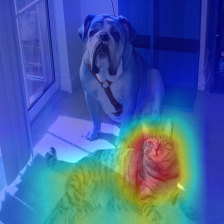}
    \end{minipage}
    \begin{minipage}[t]{0.25\linewidth}
    \centering
    \includegraphics[width=0.8in]{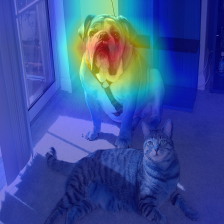}
    \end{minipage}}
    \subfigure[Tokyo image and visualization]{
    \begin{minipage}[t]{0.36\linewidth}
    \centering
    \includegraphics[width=1in]{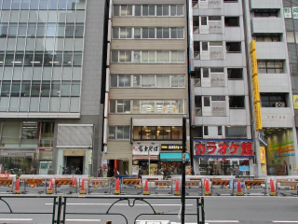}
    \end{minipage}
    \begin{minipage}[t]{0.36\linewidth}
    \centering
    \includegraphics[width=1in]{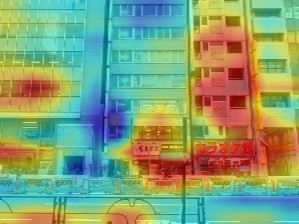}
    \end{minipage}}
    \label{fig:1}
    \caption{Visualization examples of image classification (supervised) and city recognition. (a) From left to right: original image with a cat and a dog and the visualization with 'cat'/'dog' information (highlighting cat/dog); ~\cite{selvaraju2017grad}. (b) From left to right: original image of Tokyo; visualization with 'Tokyo' information (highlighting, e.g., building, fence and signboard).}
    \vspace{-3ex}
\end{figure}

One common assumption in interpretability analysis of discriminate networks is that the  image label matches with a single dominant object. However, interpreting CNNs for city recognition deviates from this assumption as the labels of images for place recognition are places, such as names of nations or cities, which is different from discriminate objects appearing in city images such as certain architecture or vegetation type. The information on these discriminate objects is an unknown priori, including what objects and how many kinds of them are present in the data and even the same kinds of objects could appear in images of different classes. Figure~\ref{fig:1} shows an example. Obtaining the information of discriminate objects and how to interpret these visual objects in a dataset are the main stream of our study. 

This work offers a method to both qualitatively and quantitatively evaluate interpretibility of city recognition CNNs. While qualitative methods judge the interpretibility of networks directly by human ~\cite{zhou2014object,bau2017network,selvaraju2017grad}, quantitative methods compute a mathematical expression that reflects the trustworthiness. Examples of the quantitative techniques are   \cite{zhang2018interpretable,bau2017network} that compute Intersection over Union (IoU) score to evaluate the interpretability across networks as an objective confidence score. In \cite{selvaraju2017grad,fong2017interpretable} localization precision of visualizations through Pointing Game~\cite{zhang2018top} is evaluated.

To the best of our knowledge there is no work that quantitatively measures the interpretibility of CNN in a holistic manner. Previous work consider supervised visualization where the the labels of objects that are localized in the image are consistent with the class labels~\cite{zhou2016learning,selvaraju2017grad,oquab2014learning,oquab2015object}.

We raise the following research questions in this paper
and we try to address them via relevant experiments.
\renewcommand\labelitemi{$\vcenter{\hbox{\tiny$\bullet$}}$}
\setlist{nolistsep}
\begin{itemize}[noitemsep]
    \itemsep0em 
    \item Are the deep representations learned by the city recognition CNNs interpretable? 
    \item How to measure and evaluate the interpretability of in weakly supervised network?
    \item Do different architectures or initializations of CNNs affect the interpretability?
\end{itemize}

\section{Methodology}
We summarize our proposed interpretability investigations roughly in several steps:
\setlist{nolistsep}
\begin{enumerate}[noitemsep]
    \item Weighted masks are generated in the ultimate layer of any given trained CNNs model that classifies images from different cities, using Grad-CAM that highlights the class-discriminate regions of the test image. A visual explanation is generated using a threshold and weighted mask to cover unimportant regions on test image for classification.
    \item Visual explanations are visualized using t-SNE to detect meaningful patterns in an unsupervised manner.
    \item A pretrained segmentation model is used to annotate the objects in the test images pixelwise.
    \item The normalized distribution of the objects annotated in visual explanations for each class is plotted to see if there is a significant skew towards certain objects. 
\end{enumerate}

\subsection{Generating Visual Explanations}
\begin{figure}[!htb]
\centering
\vspace{-1ex}
\includegraphics[width=0.52\textwidth]{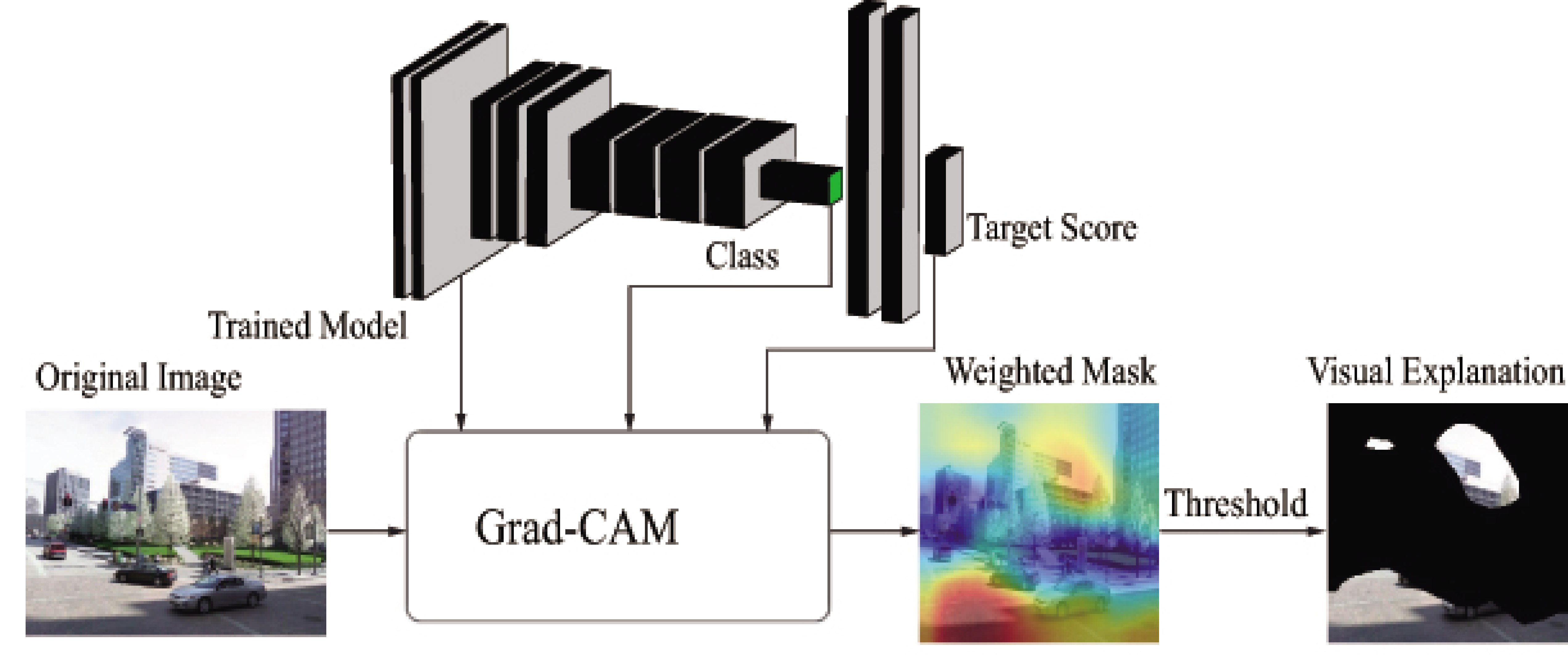}
\caption{The pipeline of generating weighted masks and visual explanations with Grad-CAM~\cite{selvaraju2017grad} for city recognition CNNs.}
\label{fig:2}
\end{figure}

We adopt Grad-CAM~\cite{selvaraju2017grad} as our visualization technique to generate visual explanations for each test image. Selvaraju et al.~\cite{selvaraju2017grad} proposed Grad-CAM based on the work of~\cite{zhou2016learning}, to map any class-discriminate activation of last convolutional layers onto input images. In the localization heat-maps ($L^c_{\textrm{Grad-CAM}}$), the values of significance are calculated in pixel level and the important regions are highlighted on input images. The localization heat-maps can be computed by a linear combination of weighted forward activation maps as proposed in~\cite{selvaraju2017grad}. Note that the weighted masks $mask\_norm$ are generated by normalizing localization heat-maps to ensure the values of significance range  between $[0,1]$ for each weighted mask. Additionally, we set a threshold to select important regions (pixels) from the weighted masks to generate visual explanations. See Figure~\ref{fig:2} for illustration.

\vspace{-3ex}
\subsection{Clustering Weighted Masks}
Due to the lack of  object labels appearing in visual explanations, we adopt unsupervised method to cluster visual explanations directly to recognize potential patterns. Proper descriptors needs to be extracted to cluster the visual explanations. Instead of extracting descriptors from visual explanations, we take the weighted masks $mask\_norm$ 
as descriptors and cluster them. 
We use t-distributed stochastic neighbor embedding (t-SNE)~\cite{maaten2008visualizing} for clustering and dimensionality reduction. 

\subsection{Quantifying Interpretability}
The aim of this study is to examine the interpretability of deep representations learned from city recognition CNNs, therefore it is necessary to obtain the information of what objects appear as discriminate in the images. In our work, we first use semantic segmentation model to label the objects in pixel level. This pretrained segmentation model should be able to recognize all classes of objects appearing in images. Hence the class information of objects can be used for evaluating the interpretability of deep representations quantitatively.

Some quantitative measurements of interpretability in previous researches, such as IoU~\cite{zhang2018interpretable,bau2017network} and Pointing Game~\cite{selvaraju2017grad,fong2017interpretable,zhang2018top}, cannot be used for city recognition CNNs, since there is inconsistency between the class information of city images and the class information of objects appearing in city images. Alternatively, we suppose objects appearing in the visual explanations are class-discriminate and their frequent occurrence reflects the interpretability of deep representations. To quantify this metric we calculate the number of pixels for different objects in visual explanations of the test images. To rule out the biases of different classes, we normalize the numbers of pixels of class-discriminate object $p$ in the visual explanations $M_{P}$ to the pixels of the same object in all images from that dataset $N_{P}$:
\begin{equation}
    R_{p}^c = \frac{\sum_{i=1}^{N^c}M_{p,i}}{\sum_{i=1}^{N^c}N_{p,i}},
\end{equation}
 where $N^c$ is the total number of city images of class $c$, indexed by $i$. For instance, $p$ can be trees where $R_{p}^c$ reflects the ratio of trees appearing as class discriminate in the class Tokyo to the whole trees appearing in this class. $R_{p}^c$ is a quantifiable bounded measure of object significance varying between $[0,1]$, where $0$ means non-discriminate with respect to other classes and $1$ means very discriminate. 

\vspace{-1ex}
\section{Experiments}
\subsection{Datasets}
We use two datasets of city images, which are \textbf{Tokyo 24/7} and \textbf{Pittsburgh} introduced from~\cite{arandjelovic2016netvlad} to obtain city recognition CNNs.
\begin{itemize}[noitemsep]
    \itemsep0em 
    \item \textbf{Tokyo 24/7}: This dataset contains 76k dataset images. For the same spot, 12 images were taken from different directions.
    \item \textbf{Pittsburgh}: This dataset contains 250k database images. For the same spot, 24 images were taken from 12 different direction and 2 different angles.
\end{itemize}

To avoid unbalanced datasets, we only use 76k Pittsburgh images. All images are divided into training, validation and test datasets with the proportions as 6:2:2. These two datasets do not contain any information of objects.

\subsection{Experimental Setup}
We train four different image classification CNNs models to classify city images. The network architectures include VGG11~\cite{simonyan2014very}, ResNet18~\cite{he2016deep} and two other shallow networks (as shown below in Table~\ref{table:1}), Simple and Simpler. These four image classification networks are used for interpreting deep representations of city recognition CNNs and investigating the influence of network architectures on the interpretability.

All four models are trained with the same training setup. The loss function is cross-entropy function, and Adam optimizer is applied. The initial learning rate is set as 0.0001 and is multiplied by 0.1 every 10 epochs. The accuracies of four models are $99.98\%$, $99.96\%$, $99.31\%$ and $98.18\%$.
\begin{table}[htb]\small
    \centering
    \caption{Configurations of two shallow networks. In this table, 'convN$\times$N' represents convolutional layer with a N$\times$N filter, and each convolutional layer is followed by a ReLU activation function. The number after hyphen represents the number of channels in the corresponding feature map, and the numbers in the brackets is the size of filter in max pooling layer.}
    \label{table:1}
    \begin{tabular}{c|c}
        \hline
        Simple & Simpler \\ \hline
        \multicolumn{2}{c}{Input images:224$\times$224$\times$3(RGB)} \\ \hline
        conv5$\times$5-20 & conv9$\times$9-20 \\ \hline
        \multicolumn{2}{c}{max pooling(2$\times$2)} \\ \hline
        conv7$\times$7-64 & conv9$\times$9-64 \\ \hline
        \multicolumn{2}{c}{max pooling(2$\times$2)} \\ \hline
        conv5$\times$5-96 & conv9$\times$9-96 \\ \hline
        \multicolumn{2}{c}{max pooling(2$\times$2)} \\ \hline
        conv7$\times$7-128 &  \\ \hline
        max pooling(2$\times$2) &  \\ \hline
        \multicolumn{2}{c}{fully connected-4096} \\ \hline
        \multicolumn{2}{c}{fully connected-100} \\ \hline
        \multicolumn{2}{c}{fully connected-number of classes:2} \\ \hline
    \end{tabular}
\end{table}

\subsection{Clustering Weighted masks}
\begin{figure}[!htb]
\centering
\vspace{-2.5ex}
    \includegraphics[width=0.4\textwidth]{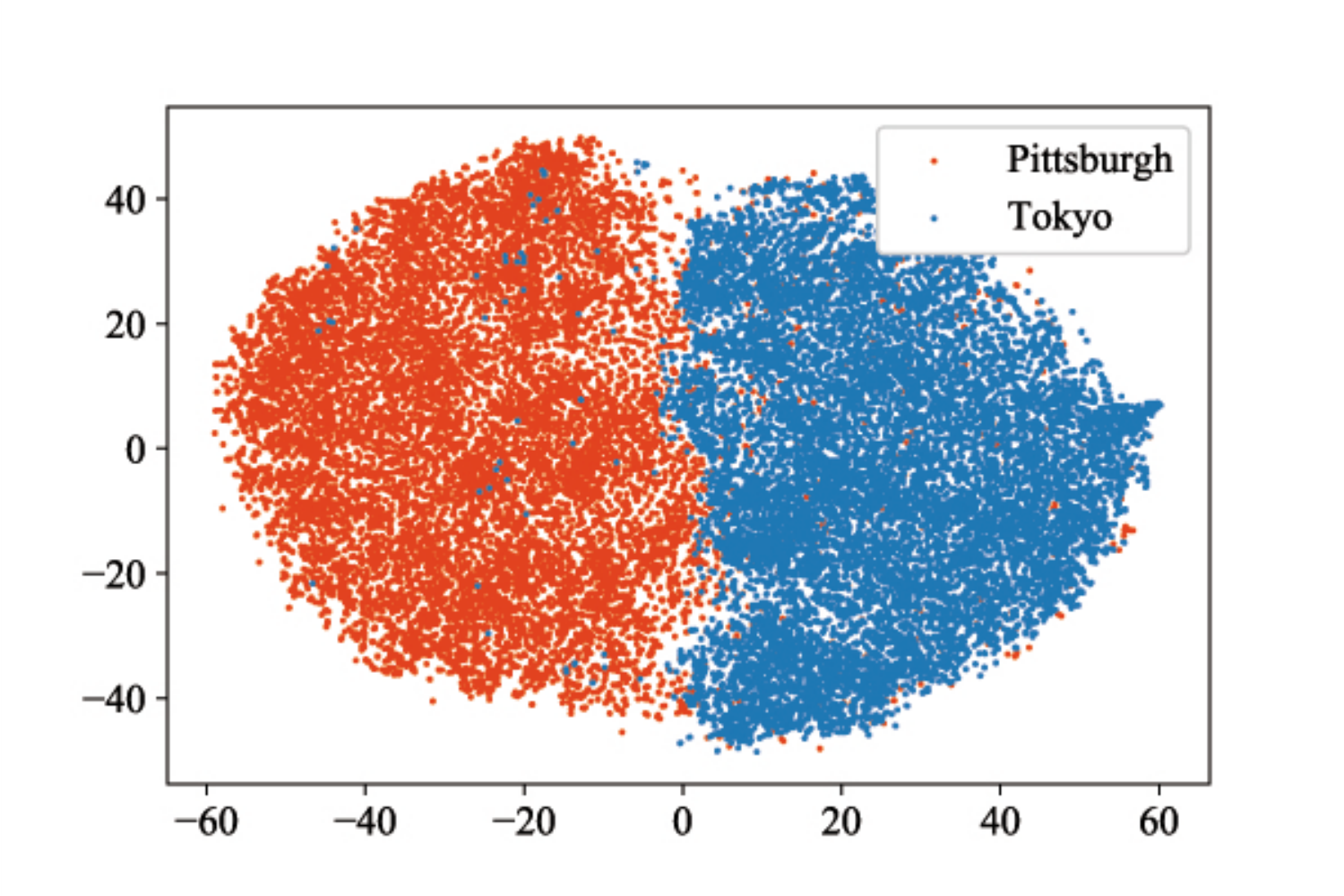}
\caption{Scatter plots of clustering results of VGG11 with city information of images. Each point represents a weighted mask generated from each test image. Most of weighted masks from different datasets are separable in terms of city label information.}
\vspace{-2ex}
\label{fig:3}
\end{figure}

To address our first research question on whether the learned representation in the last convolutional layer of our trained CNN are interpratable by human or not, we conduct the following experiment. Using t-SNE, the weighted masks ($mask\_norm$) are clustered in an unsupervised manner instead of visual explanations due to the lack of objec-level labels and the irregular shapes of black regions around visual explanations. We apply PCA to extract 50 dominating features prior the the t-SNE clustering and dimensionality reduction. Figure~\ref{fig:3} shows the scatter plots of VGG11 clustering results with label information of city images.

\begin{figure}[!htb]
    \centering
    \includegraphics[width=0.43\textwidth]{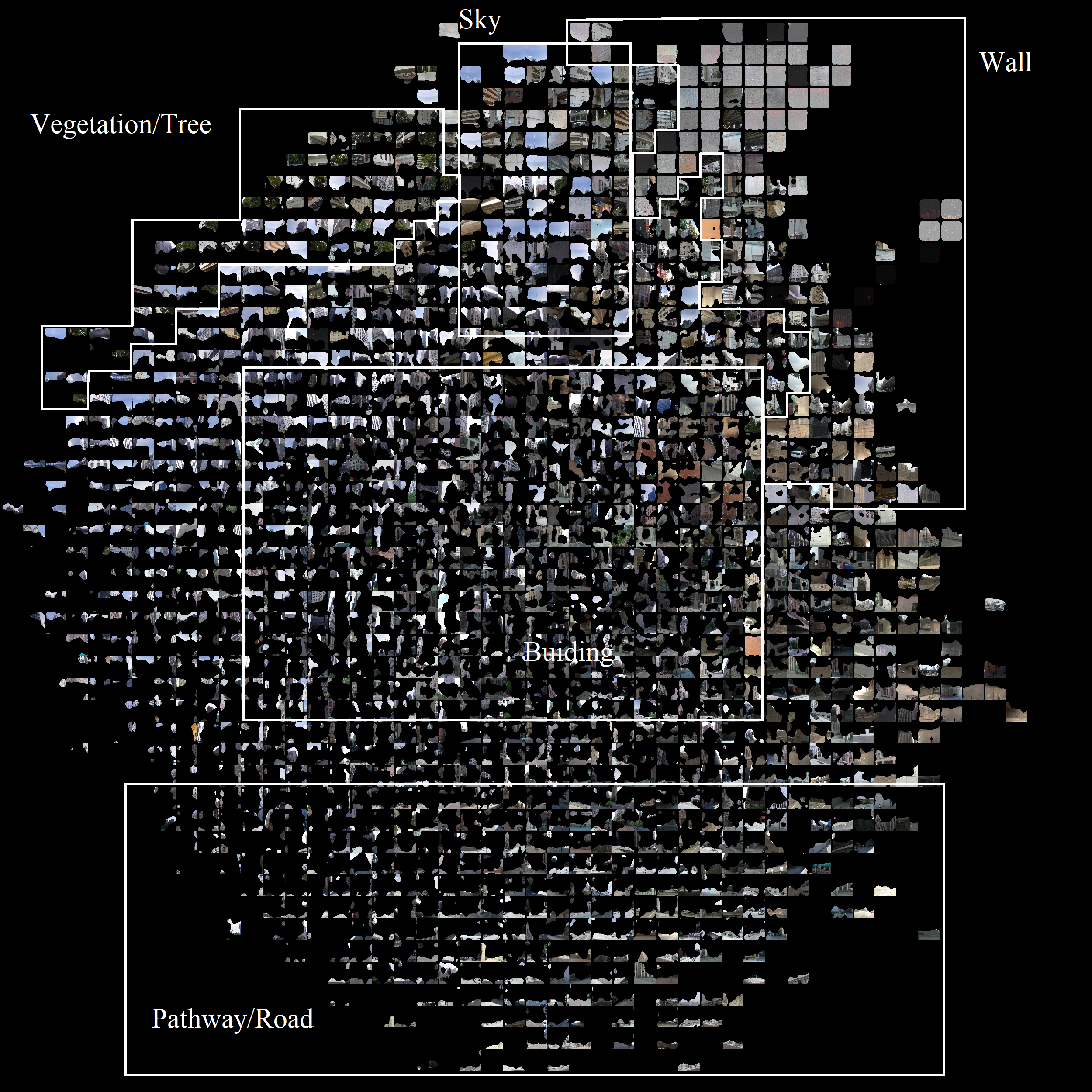}
    \caption{Exhibiting t-SNE results with visual explanations of VGG11. After replacing clustering reslut with visual explanations of test images, similar class-discriminate objects in visual explanations are clustered together. The information of these objects is obtained directly by human.}
    \label{fig:4}
    \vspace{-2ex}
\end{figure}

From the Figure~\ref{fig:3}, the clustering result that the weighted masks of test images are separable, is consistent with the high accuracy of VGG11. To visually exhibit the objects information in visual explanations that is related with the interpretability of deep representations, we next replace the points with visual explanations and demonstrate the relation between clustering result and class-discriminate objects intuitively. Due to the considerable number of test images, we randomly select around 500 visual explanations generated from VGG11 model to exhibit, as shown in Figure~\ref{fig:4}.

Based on the data visualization results shown in Figure~\ref{fig:4}, we can see that the result of our clustering leads to a collection of visually similar objects in a 2D map, which indicates that the VGG11 model learns semantically meaningful  discriminate objects in the last convolutional layer. Although these patterns of objects reveal the interpretability of deep representations learned for a city recognition CNN, to some degree, it is still necessary to evaluate the interpretability in a quantitative manner. 

\subsection{Object-level Interpretability}
We address the second research question in this section by quantifying the object level information that are extracted using visualization method. 
The lack of classes information of objects appearing in city images from \textbf{Pittsburgh} and \textbf{Tokyo 24/7} datasets makes it difficult to quantify the interpretibility of the deep representations learned from a city recognition CNN. Therefore, we apply semantic segmentation models to obtain the objects classes information before evaluating interpretability. The semantic segmentation model used in our experiment is pre-trained on MIT ADE20K scene parsing dataset~\cite{zhou2016semantic,zhou2017scene,xiao2018unified} and is built on ResNet50~\cite{he2016deep}. The segmentation model is able to classify 150 different categories of objects, including all classes of objects appearing in city images.
\begin{figure}[!htb]
    \centering
    \vspace{-1ex}
    \subfigure[Histogram of $R_{p}^c$ over class-discriminate objects appearing in \textbf{Pittsburgh}]{
    \centering
    \includegraphics[width=0.47\textwidth]{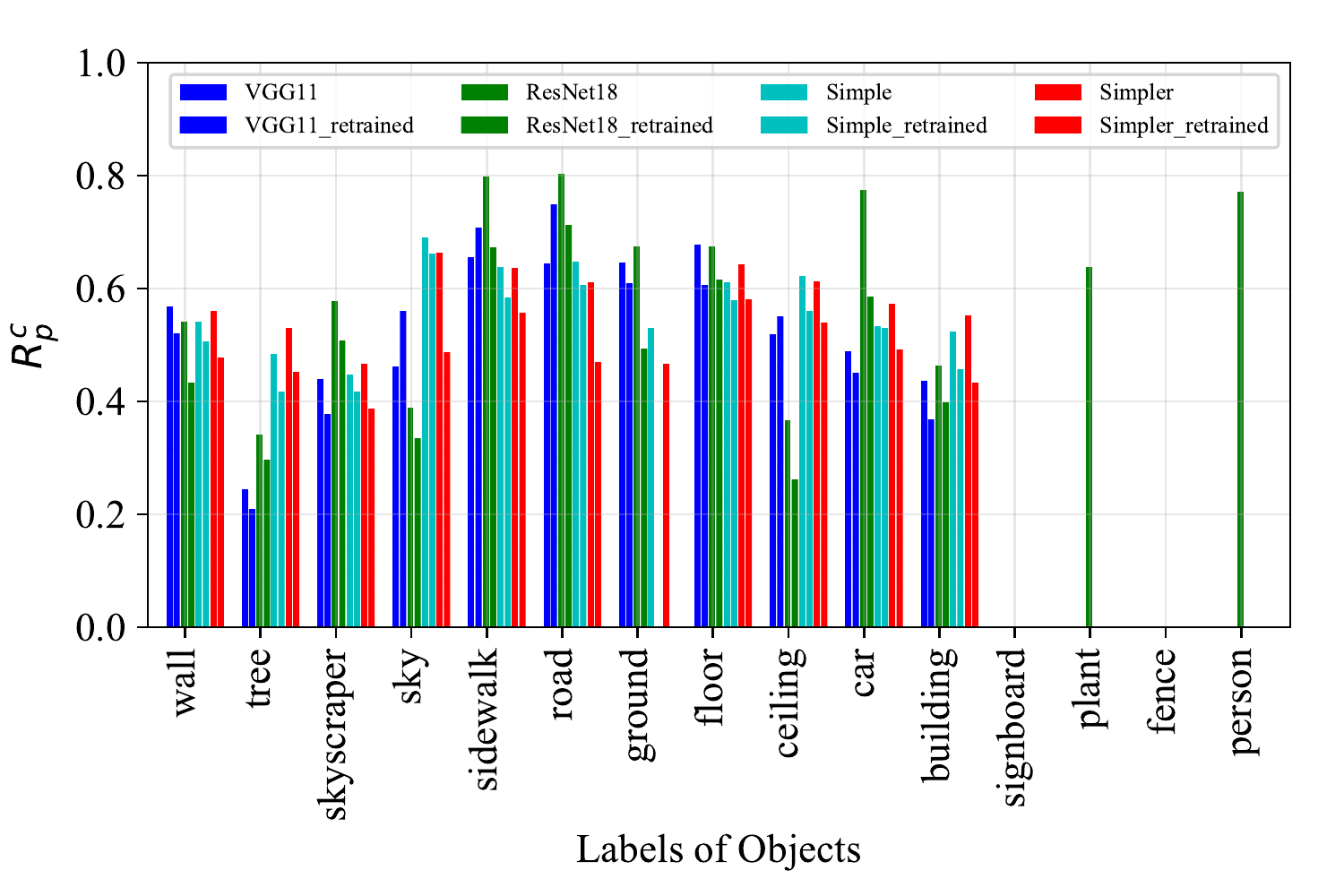}
    \vspace{-2ex}}
    \subfigure[Histogram of $R_{p}^c$ over class-discriminate objects appearing in \textbf{Tokyo 24/7}]{
    \centering
    \includegraphics[width=0.47\textwidth]{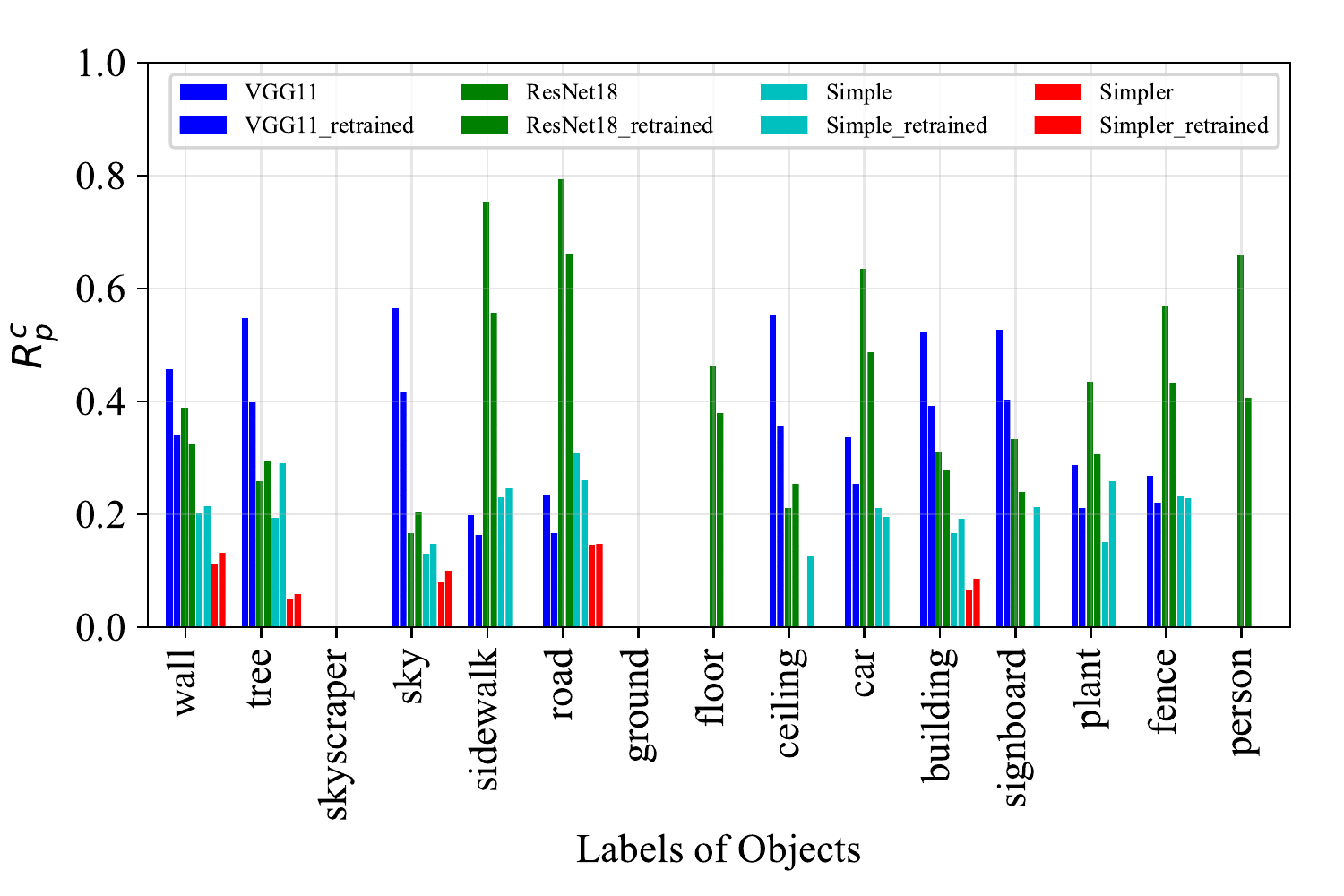}}
    \caption{Histograms of $R_{p}^c$ regarding different architectures of CNNs, initializations and datasets. Different values of $R_{p}^c$ of different objects are learned from different datasets. Some unique objects can only be learned from certain dataset.}
    \label{fig:histogram}
    \vspace{-3ex}
\end{figure}

To evaluate interpretability of deep representations quantitatively and avoid missing any possible information of objects in visual explanations, we calculate $R_{p}^c$ for different objects and datasets (classes), as shown in Figure~\ref{fig:histogram}. The objects are selected by the criterion that the average number of pixels exceeds a certain threshold (set as 100). 

Comparing the class-discriminate objects shown in Figure~\ref{fig:histogram}, dissimilar objects for different datasets are learned by city recognition CNNs. Skycraper and ground are the unique class-discriminate objects learned from \textbf{Pittsburgh} dataset, while signboard and fence are the unique ones from \textbf{Tokyo 24/7}. The values of the ratios of pixels $R_{p}^c$ indicate the selectivity of city recognition CNNs from specific dataset. The larger value of $R_{p}^c$ is, the stronger the class-discriminate attributes. E.g., a uniform histogram over different objects means city recognition CNNs take any object in the image as class-discriminate, which is meaningless in this case. Different non-uniform distributions over objects from different classes reveal city recognition CNNs learn distinct combinations of class-discriminate objects from different datasets, which is interpretable for city recognition CNNs.  

\subsubsection{Do Different Models Learn Similar discriminate Objects?}
\begin{figure}[!htb]
\vspace{-1ex}
\centering
\includegraphics[width=0.45\textwidth]{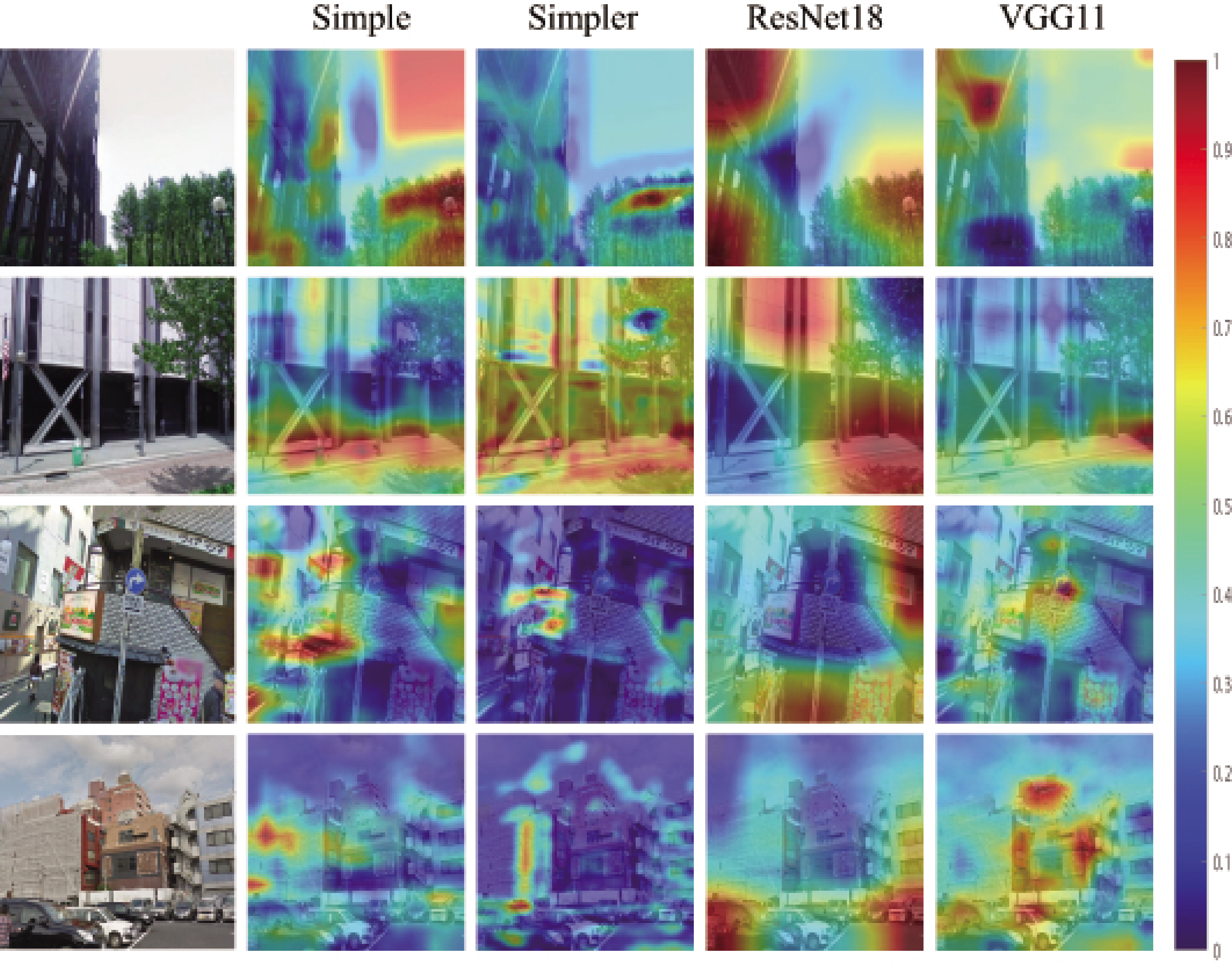}
\caption{Different city recognition CNNs generate different weighted masks for the same image. The first two rows present two test images from \textbf{Pittsburgh} and the last two rows show the test images from \textbf{Tokyo 24/7}. From the second column to the fifth column, the weighted masks are shown for Simple, Simpler, ResNet18 and VGG11 city recognition CNNs, respectively. Note that a shallow net triggers on the sky or on disjoint regions in the image. The ResNet18 focuses on wider regions and VGG11 is more selective.}
\label{fig:different}
\end{figure}

Besides the histograms used in Figure~\ref{fig:histogram}, we also apply another quantitative method to investigate the influence of network architectures and initializations on the interpretability of city recognition CNNs. Figure~\ref{fig:different} shows some examples of weighted masks learned by different city recognition CNNs. The difference among the weighted masks learned by different city recognition models reflects the influence of network architectures.

To quantify the divergence between different models, we calculate the average residual $AR$ of each city image between any two models to investigate the consistency quantitatively:
\begin{equation}
    AR_{m_1,m_2} = \frac{\left|mask\_norm^c_{m_1}-mask\_norm^c_{m_2}\right|}{H\times W},
\end{equation}
where $H$ and $W$ are the height and width of weighted masks, and $m_1$ and $m_2$ represent CNN models. The value of $AR_{m_1,m_2}$ ranges from 0 to 1, where 0 means two models learn exactly same weighted mask for this image and 1 means totally different weighted masks have been learned by two models. Due to the considerable images, we calculate the average $AR_{m_1,m_2}$ over all test images. All values of average $AR_{m_1,m_2}$ between different network architectures and initialziations are listed in Table~\ref{ar}.

Comparing the values in Table~\ref{ar}, we can find that $AR$s between different architectures are larger than the ones between different initializations in general, which means network architectures affect the deep representations learned from city recognition CNNs greater than different training initializations. This is also consistent with the results from Figure~\ref{fig:histogram}.
\begin{table}[htb]\footnotesize	
\centering
\caption{The average $AR_{m_1,m_2}$ between different network architectures and initializations. The values of $AR_{m_1,m_2}$ between different network architectures are all larger than the ones between different initializations.}
\label{ar}
\begin{tabular}{c|c}
\hline
Models ($m_1$-$m_2$) & Average $AR_{m_1,m_2}$ \\ \hline
VGG11-ResNet18 & 0.4349 \\ \hline
VGG11-Simple & 0.4303 \\ \hline
VGG11-Simpler & 0.4502 \\ \hline
ResNet18-Simple & 0.4118 \\ \hline
ResNet18-Simpler & 0.4136 \\ \hline
Simple-Simpler & 0.3149 \\ \hline
VGG11-VGG11\_retrained & 0.2679 \\ \hline
ResNet18-ResNet18\_retrained & 0.2265 \\ \hline
Simple-Simple\_retrained & 0.2411 \\ \hline
Simpler-Simpler\_retrained & 0.2460 \\ \hline
\end{tabular}
\end{table}

Besides calculating $AR$s between different city recognition models, we can also use $R_{p}^c$ to get the consistent results. In Figure~\ref{fig:histogram} (a) and (b), different CNNs architectures learn dissimilar histograms over class-discriminate objects, however, similar values of $R_{p}^c$ over class-discriminate objects are learned due to different initializations. In Figure~\ref{fig:histogram} (b), we can also find the values $R_{p}^c$ of VGG11 and ResNet18 are larger than the ones of shallow networks over all class-discriminate objects, which also reflects that convolutional features learned by deep network architectures are more semantically interpretative than the shallow ones. Therefore, the influence of network architectures on the interpretability of CNN features is stronger than the one of different initializations.

\vspace{-1ex}
\section{Conclusion}
In this work, we provided a framework to investigate the emergence of semantic objects as discriminate attributes in the ultimate layer of network.  This is consistent with the way human understand city images. We applied our proposed framework to investigate the influence of network architectures and different initializations on the interpretability. We conclude that network architectures would affect the learned visual representations greater than different initializations.
{\small
\bibliographystyle{ieee}
\bibliography{egpaperfinal}
}
\end{document}